# Edge-aware GAT-based protein binding site prediction

Weisen Yang[1], Hanqing Zhang[1], Wangren Qiu[1], Xuan Xiao[1,2], Weizhong Lin[1]

1. *School of Information Engineering, Jingdezhen Ceramic University, Jingdezhen, China, 333403*

2. *School of Information Engineering, Jiangxi Art & Ceramics Technology Institute, Jingdezhen, 333000, China.*

## Abstract

Accurate identification of protein binding sites is crucial for understanding biomolecular interaction mechanisms and for the rational design of drug targets. Traditional predictive methods often struggle to balance prediction accuracy with computational efficiency when capturing complex spatial conformations. To address this challenge, we propose an Edge-aware Graph Attention Network (Edge-aware GAT) model for the fine-grained prediction of binding sites across various biomolecules, including proteins, DNA/RNA, ions, ligands, and lipids. Our method constructs atom-level graphs and integrates multidimensional structural features, including geometric descriptors, DSSP-derived secondary structure, and relative solvent accessibility (RSA), to generate spatially aware embedding vectors. By incorporating interatomic distances and directional vectors as edge features within the attention mechanism, the model significantly enhances its representation capacity. On benchmark datasets, our model achieves a ROC-AUC of 0.93 for protein-protein binding site prediction, outperforming several state-of-the-art methods. The use of directional tensor propagation and residue-level attention pooling further improves both binding site localization and the capture of local structural details. Visualizations using PyMOL confirm the model's practical utility and interpretability. To facilitate community access and application, we have deployed a publicly accessible web server at http://119.45.201.89:5000/. In summary, our approach offers a novel and efficient solution that balances prediction accuracy, generalization, and interpretability for identifying functional sites in proteins.

**Keywords:** Protein binding site prediction, Graph Attention Network, Deep learning, Structural bioinformatics, Edge-aware attention, Molecular interactions

# Introduction

Proteins are fundamental components of life, regulating most biological functions through their specific molecular interactions. In biological systems, molecular interfaces are ubiquitous, playing a central role in the formation of cellular boundaries and intracellular organization. However, their functional significance extends beyond these roles. Proteins exert their biological functions by interacting with other proteins, nucleic acids, cell membranes, various small molecules, and ions. The analysis and comparison of protein binding sites are critical for a wide range of applications in drug discovery, including de novo drug identification, drug repurposing, and polypharmacology [1]. Understanding the structure and properties of proteins—especially their binding sites—is essential for elucidating their biological functions and for developing therapeutic agents. Computational techniques have become indispensable in the drug discovery process, enabling approaches such as virtual screening of small molecules [2], structure-based drug design, and molecular docking between small molecules and proteins. Studies analyzing ligand-protein complexes in the Protein Data Bank (PDB) [3] have shown that most ligands interact with specific binding sites on their target proteins. Each of these binding sites is characterized by a unique set of properties or biological functions that distinguish them from other surface cavities on the protein [4].

Predicting whether a given protein can interact with other molecules remains a significant challenge in biology. Despite substantial progress in various areas, this problem is still far from being fully solved [5]. Many machine learning-based methods have been developed for protein interaction site prediction [6-17], further expanding the methodological options in this field. These earlier studies laid the foundation for many current approaches, exploring various machine learning algorithms and feature extraction techniques. However, the field has advanced considerably since then, with newer methods incorporating deep learning and sophisticated structural analysis to achieve greater accuracy and efficiency.

The emergence of deep learning techniques has led to the development of several effective approaches for protein binding site prediction. These include methods based on spherical convolutional neural networks such as Spherical CNN [18], DeepSphere [19], geometric deep learning-based frameworks like MaSIF, which predicts molecular surface interactions [20], structure-based predictors such as ScanNet [21]、and parameter-free geometric learning models like PeSTo, which target protein interaction interfaces [22]. These methods have demonstrated promising results across diverse large-scale datasets. However, despite their importance, such experimental and computational techniques often come with high costs, significant computational complexity, long runtimes, and complicated operational procedures. Specifically, computing protein surface representations and extracting surface features are time-consuming and technically challenging. Furthermore, surface parameterization is usually sensitive to structural details and errors in protein models [23]. Therefore, developing an efficient and robust computational framework for predicting protein binding sites remains an urgent and valuable task.

Early representative models such as DeepSite [24], CNNSite [25], CNN-LSTM [26], and Frsite [27] demonstrated that CNNs and LSTMs could effectively learn complex protein patterns. In recent years, deep learning techniques—particularly those that incorporate structural information—have demonstrated tremendous potential in modeling molecular interactions within the field of bioinformatics. With the emergence of Graph Convolutional Networks (GCNs) in structural analysis of proteins, researchers began to leverage these advanced neural architectures to process three-dimensional protein structural data. Graph Neural Networks (GNNs) [28] are particularly well-suited for modeling the structural and interactional properties of proteins, as amino acid chains often exhibit complex spatial configurations that naturally form graph-like structures. By applying graph convolution, it is possible to effectively capture both local structural features and global topological information, thereby significantly improving the accuracy of binding site prediction. Zitnik et al. demonstrated the use of GNNs for modeling protein interactions and structural representations with promising results [29]. Building on this, Ryu developed an optimized GNN-based model for protein-ligand binding prediction, enhancing both accuracy and computational efficiency [30]. Furthermore, studies by Yuan [31] and Zaki [32] integrated protein spatial structures into GNN frameworks to further improve the prediction of functional sites. Compared to traditional Convolutional Neural Networks (CNNs), GCNs are inherently more capable of handling non-Euclidean data structures, making them especially advantageous for modeling complex biomolecules like proteins with intricate spatial relationships.

Many successful approaches for protein binding site prediction have combined Transformer architectures with geometric deep learning [20], representing protein structures as graphs or point clouds. These methods leverage the translation-invariance properties of neural networks to model protein geometry and spatial interactions effectively [33-35]. Building on these principles, two mainstream paradigms have emerged: voxel-based and graph-based representations, each with distinct computational strategies. Voxel-based methods discretize protein structures into 3D grids, enabling direct application of Conventional Neural Networks (CNNs). OctSurf [36] and 3D U-Net [37] pioneered this approach by utilizing CNN to voxel grids for simultaneous detection of surface and interior binding sites. Similarly, Pinheiro [38] proposed a voxel-based CNN for drug design, while VoxPred [39] enhanced the paradigm by combining raw 3D structural data with voxel grids, achieving improved predictive performance across multiple datasets. Collectively, these models demonstrate the robustness of CNN on regular grids, though they share trade-offs in resolution and computational cost. Graph-based approaches, by contrast, operate directly on irregular molecular structures, preserving topological information. Son [40] employed GCNs to explicitly model amino acid interactions, achieving significant accuracy gains by capturing spatial adjacency more naturally than voxel discretization. Extending geometric deep learning beyond Euclidean grids, Igashov [41] introduced Spherical CNNs to process spherical protein surface data, offering a rotation-equivariant alternative that improves robustness and precision in surface feature and binding site prediction.

This progression from voxel grids to graphs and spherical representations reflects a broader trend: moving from brute-force discretization toward structure-aware, geometrically principled architectures that better respect protein. However, most existing methods based on GCNs suffer from two major limitations. First, many approaches fail to effectively incorporate spatial directional information, making them inadequate for capturing the complex anisotropic interactions within molecular structures. Second, most models operate on coarse-grained residue-level features, which limits their ability to capture fine-grained atomic-level interactions and thereby constrains prediction accuracy and generalization performance. In this study, we propose an edge-aware GAT-based protein binding site predictor. The model constructs graphs at the atomic level and integrates both node (atom-level) features and edge features to enable efficient encoding and propagation of fine-grained spatial information.

## Result

**Performance Evaluation on Multiple Binding Site Prediction Tasks**

To comprehensively evaluate our Edge-aware Graph Attention Network model, we employed four standard evaluation metrics: Accuracy, F1-score, Matthews Correlation Coefficient (MCC), and ROC-AUC. The model was assessed across five distinct molecular interaction categories: protein-protein, DNA/RNA, ion, ligand, and lipid binding sites. The quantitative results are summarized in Table 1.

Table 1. Comprehensive performance evaluation of the Edge-aware GAT model across different binding site categories.

| Binding Category | Accuracy | F1-score | MCC | ROC-AUC |
| --- | --- | --- | --- | --- |
| Protein | 0.933 | 0.771 | 0.677 | 0.930 |
| DNA/RNA | 0.911 | 0.512 | 0.525 | 0.933 |
| Ion | 0.872 | 0.449 | 0.464 | 0.841 |
| Ligand | 0.927 | 0.501 | 0.361 | 0.830 |
| Lipid | 0.736 | 0.323 | 0.459 | 0.921 |

**Visual Analysis of Predicted Binding Sites**

To further validate the model's capability in identifying protein binding sites, we conducted extensive visual analysis across five molecular interaction categories. The predicted high-probability binding regions were visualized using heatmaps overlaid on protein tertiary structures. Specifically, we encoded each atom's predicted binding probability into the B-factor field of the corresponding PDB file, enabling 3D heatmap visualization of potential interaction sites using PyMOL, where red indicates high binding probability and green indicates low probability.

For enhanced biological interpretability, we aggregated atoms with predicted binding probabilities above a threshold of 0.5 into residue-level binding sites and exported the results as CSV files. Each file includes chain ID, residue index, residue name, and mean binding probability, facilitating downstream functional annotation and comparative analysis.

**Case Studies of Representative Protein Complexes**

To demonstrate the model's predictive capability and spatial localization accuracy, we conducted a series of detailed case studies on representative protein complexes across different interaction categories. For each case, the predicted binding probabilities were encoded into the B-factor field of the corresponding PDB file, enabling the visualization of high-probability interaction sites as 3D heatmaps using PyMOL. In these visualizations, red regions indicate a high binding probability, while green regions denote a low probability, providing an intuitive assessment of the predicted interfaces. The following analyses present specific predictions for protein-protein, protein-nucleic acid, ion, ligand, and lipid binding sites, comparing model outputs with known structural data.

**Protein-Protein Interaction:** As shown in Figure 1 and Table 2, the binding site prediction for PDB structure 1DZL_A demonstrates the model's precision in identifying interfacial residues. The 3D heatmap shows concentrated red regions at the known interaction interface. Correspondingly, the model successfully identified key interfacial residues with high confidence, including ARG41 (0.9459), LEU61 (0.9391), and LEU43 (0.9341).

**Protein-Nucleic Acid Interaction:** Figure 2 and Table 3 illustrate the prediction for nucleic acid binding sites in PDB structure 1H9D_A. The model assigned high probabilities to residues critical for nucleic binding, such as LEU71 (0.9945), PHE70 (0.9826), and ASP66 (0.9558), aligning with the expected interaction region.

**Ion Binding Sites:** For PDB structure 4TSY_B, the predicted ion binding sites are visualized in Figure 3 and Table 4. The model localized the binding region, identifying residues like GLY27 (0.8402), VAL29 (0.8171), and LEU14 (0.7555) as the key coordination site.

**Ligand Binding Sites:** Figure 4 and Table 5 present the ligand binding site prediction for PDB structure 5B3Z_A. The heatmap clearly demarcates the binding pocket, and the model pinpointed central interacting residues, notably ARG18 (0.9819), GLU32 (0.9727), and ASN23 (0.9714).

**Lipid Binding Sites:** Finally, the lipid binding site prediction for PDB structure 6NFU_C is shown in Figure 5 and Table 6. The visualization reveals a surface patch likely involved in membrane interaction, with high-probability residues including LEU24 (0.9620), ARG27 (0.9403), and VAL34 (0.9257).

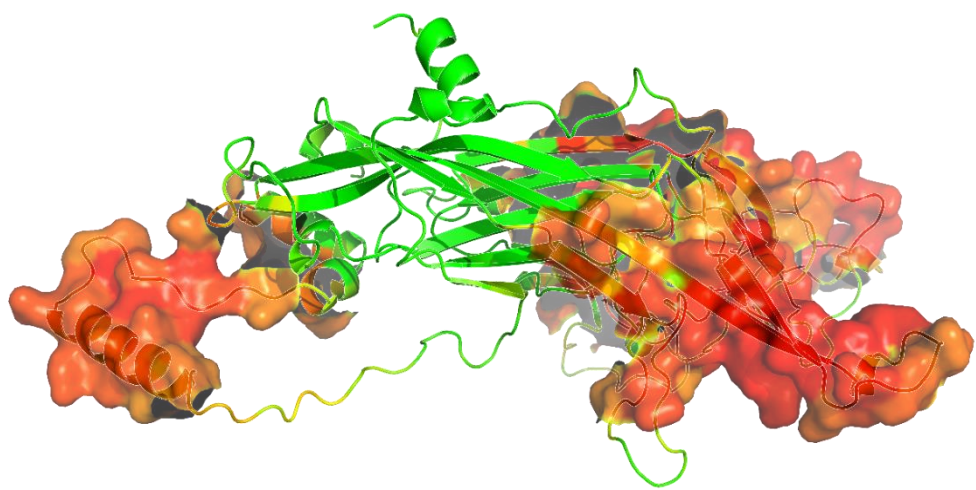

**Figure 1.** Visualization of predicted protein-protein binding sites for PDB structure 1DZL_A. Red regions indicate high binding probability.

Table 2. High-confidence protein binding sites predicted for 1DZL_A

| chain | residue_id | residue_name | mean_probability |
|-------|------------|--------------|------------------|
| A | 41 | ARG | 0.9459 |
| A | 61 | LEU | 0.9391 |
| A | 43 | LEU | 0.9341 |
| A | 23 | SER | 0.9218 |
| A | 64 | LYS | 0.9190 |
| A | 49 | TYR | 0.9172 |
| A | 60 | ILE | 0.9162 |
| A | 63 | PRO | 0.9053 |
| A | 34 | TYR | 0.8958 |
| A | 53 | LYS | 0.8925 |
| A | 32 | ASN | 0.8895 |
| A | 36 | HIS | 0.8864 |

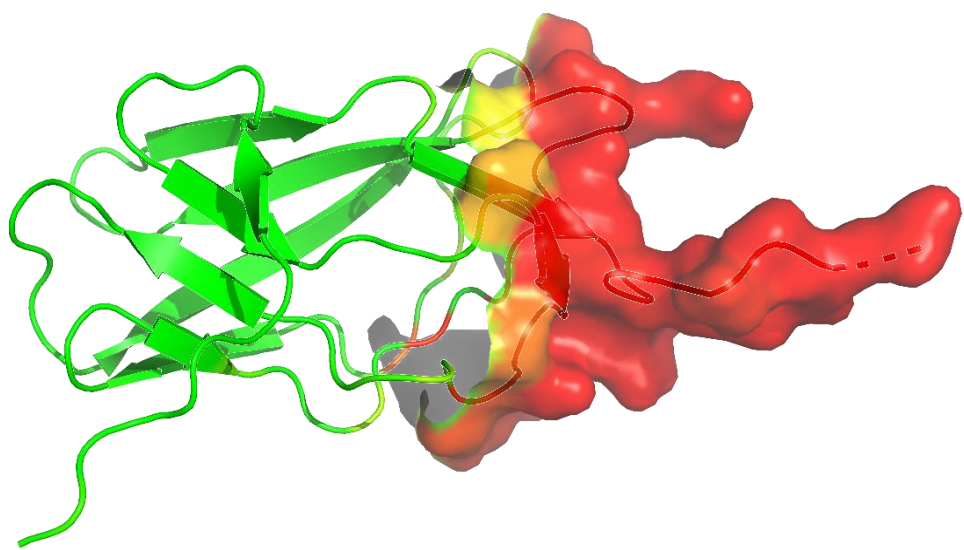

**Figure 2**. Visualization of nucleic acid binding sites for PDB structure 1H9D_A.

Table 3. Predicted nucleic acid binding sites for 1H9D_A

| chain | residue_id | residue_name | mean_probability |
|---|---|---|---|
| A | 71 | LEU | 0.9945 |
| A | 70 | PHE | 0.9826 |
| A | 66 | ASP | 0.9558 |
| A | 62 | LEU | 0.9558 |
| A | 65 | THR | 0.9354 |
| A | 58 | HIS | 0.9119 |
| A | 59 | PRO | 0.8847 |
| A | 63 | VAL | 0.8224 |
| A | 67 | SER | 0.5176 |

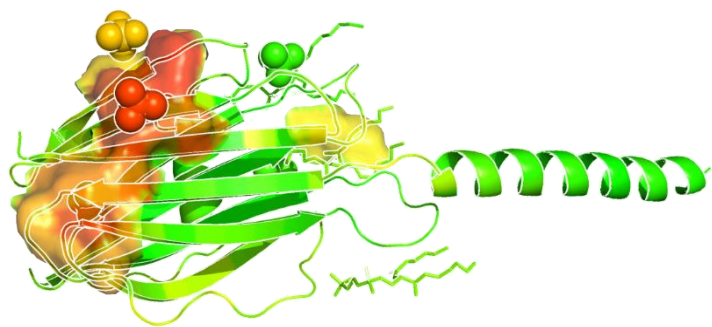

**Figure 3.** Visualization of ion binding sites for PDB structure 4TSY_B.

**Table 4.** Predicted ion binding sites for 4TSY_B

| chain | residue_id | residue_name | mean_probability |
|-------|------------|--------------|------------------|
| B | 27 | GLY | 0.8402 |
| B | 29 | VAL | 0.8171 |
| B | 14 | LEU | 0.7555 |
| B | 9 | ILE | 0.7514 |
| B | 30 | LYS | 0.7398 |
| B | 10 | ASP | 0.7199 |
| B | 25 | ALA | 0.6785 |
| B | 15 | GLY | 0.6730 |
| B | 28 | ASN | 0.5241 |

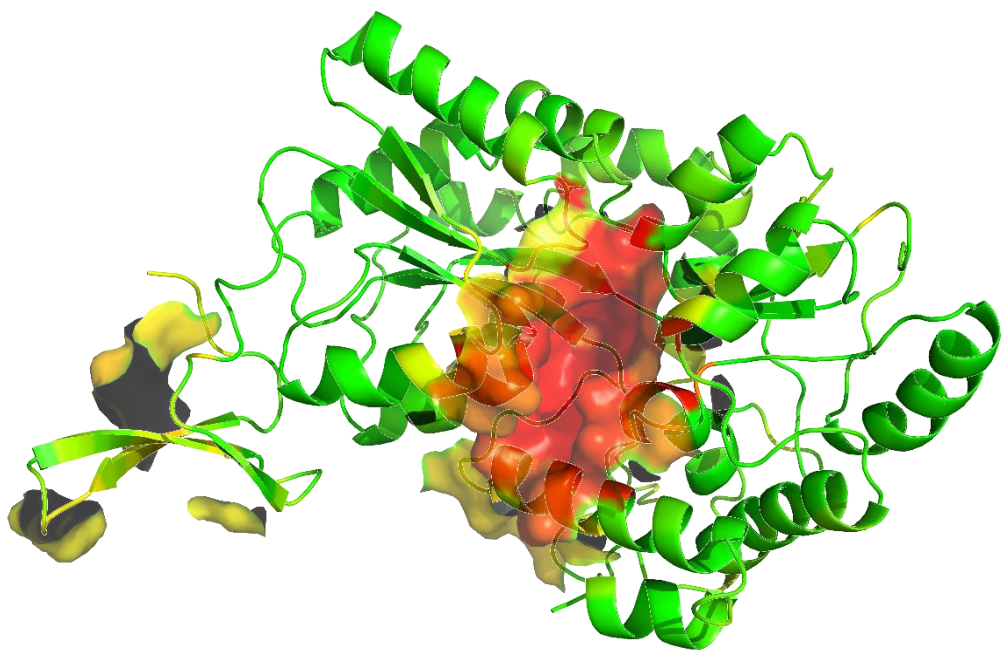

**Figure 4.** Visualization of ligand binding sites for PDB structure 5B3Z_A.

**Table 5.** Predicted ligand binding sites for 5B3Z_A

| chain | residue_id | residue_name | mean_probability |
|-------|------------|--------------|------------------|
| A | 18 | ARG | 0.9819 |
| A | 32 | GLU | 0.9727 |
| A | 23 | ASN | 0.9714 |
| A | 36 | GLY | 0.9350 |

| chain | residue_id | residue_name | mean_probability |
|---|---|---|---|
| A | 6 | PRO | 0.9116 |
| A | 30 | GLN | 0.8912 |
| A | 12 | MET | 0.8876 |
| A | 35 | SER | 0.8327 |
| A | 7 | GLY | 0.8275 |
| A | 46 | TRP | 0.8219 |
| A | 22 | PHE | 0.8076 |
| A | 10 | LYS | 0.7337 |
| A | 45 | ILE | 0.6961 |
| A | 31 | TRP | 0.5868 |
| A | 25 | ILE | 0.5774 |

**Figure 5.** Visualization of lipid binding sites for PDB structure 6NFU_C.

**Table 6.** Predicted lipid binding sites for 6NFU_C.

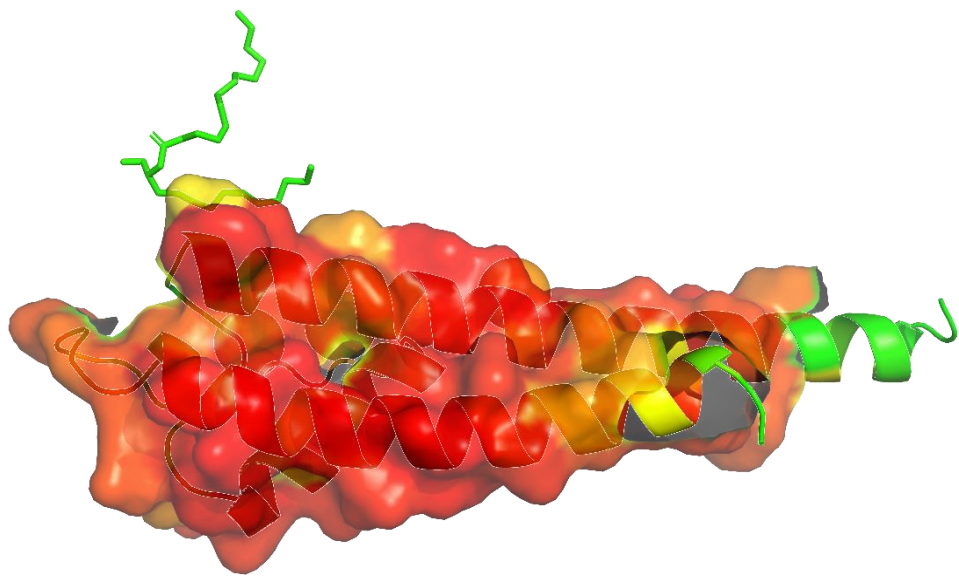

| chain | residue_id | residue_name | mean_probability |
|---|---|---|---|
| C | 24 | LEU | 0.9620 |
| C | 27 | ARG | 0.9403 |
| C | 34 | VAL | 0.9257 |
| C | 33 | THR | 0.8998 |
| C | 35 | LEU | 0.8976 |
| C | 28 | ALA | 0.8802 |
| C | 25 | HIS | 0.8796 |
| C | 26 | TRP | 0.8574 |
| C | 22 | SER | 0.8301 |
| C | 30 | GLY | 0.8140 |
| C | 29 | ALA | 0.8096 |
| C | 32 | ALA | 0.7953 |

**Web-Based Prediction System Implementation**

To make our Edge-aware GAT model accessible to the broader research community, we developed an interactive web-based prediction platform (Figure 6), publicly accessible at http://119.45.201.89:5000/. The system is designed to provide researchers with an efficient and user-friendly tool for predicting protein binding sites across five molecular categories: protein-protein, DNA/RNA, ion, ligand, and lipid interactions.

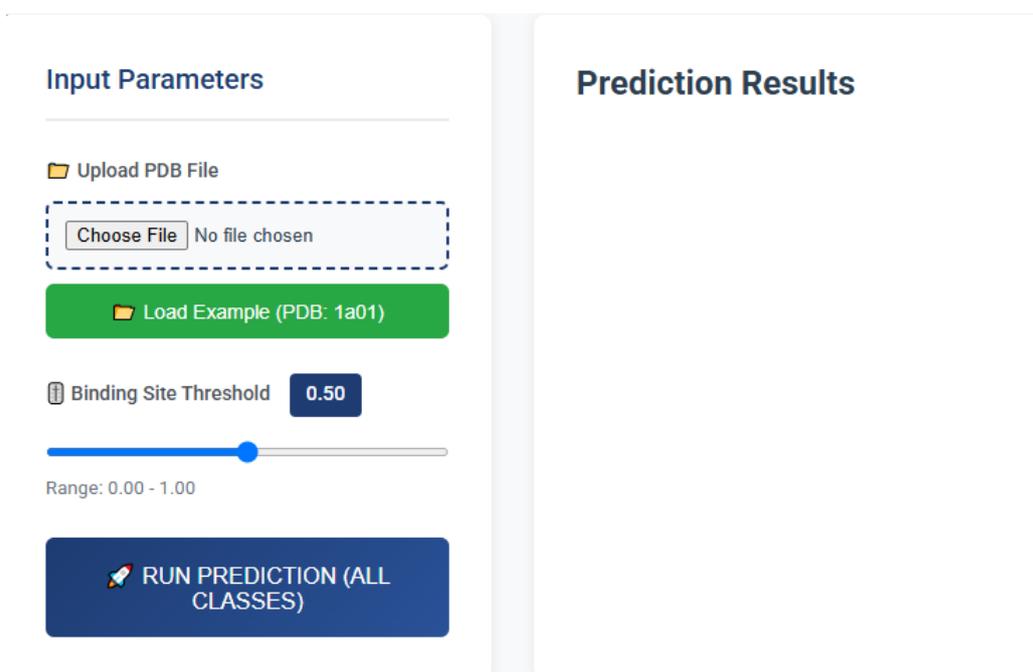

**Figure 6**. Web interface of the Edge-aware GAT prediction server

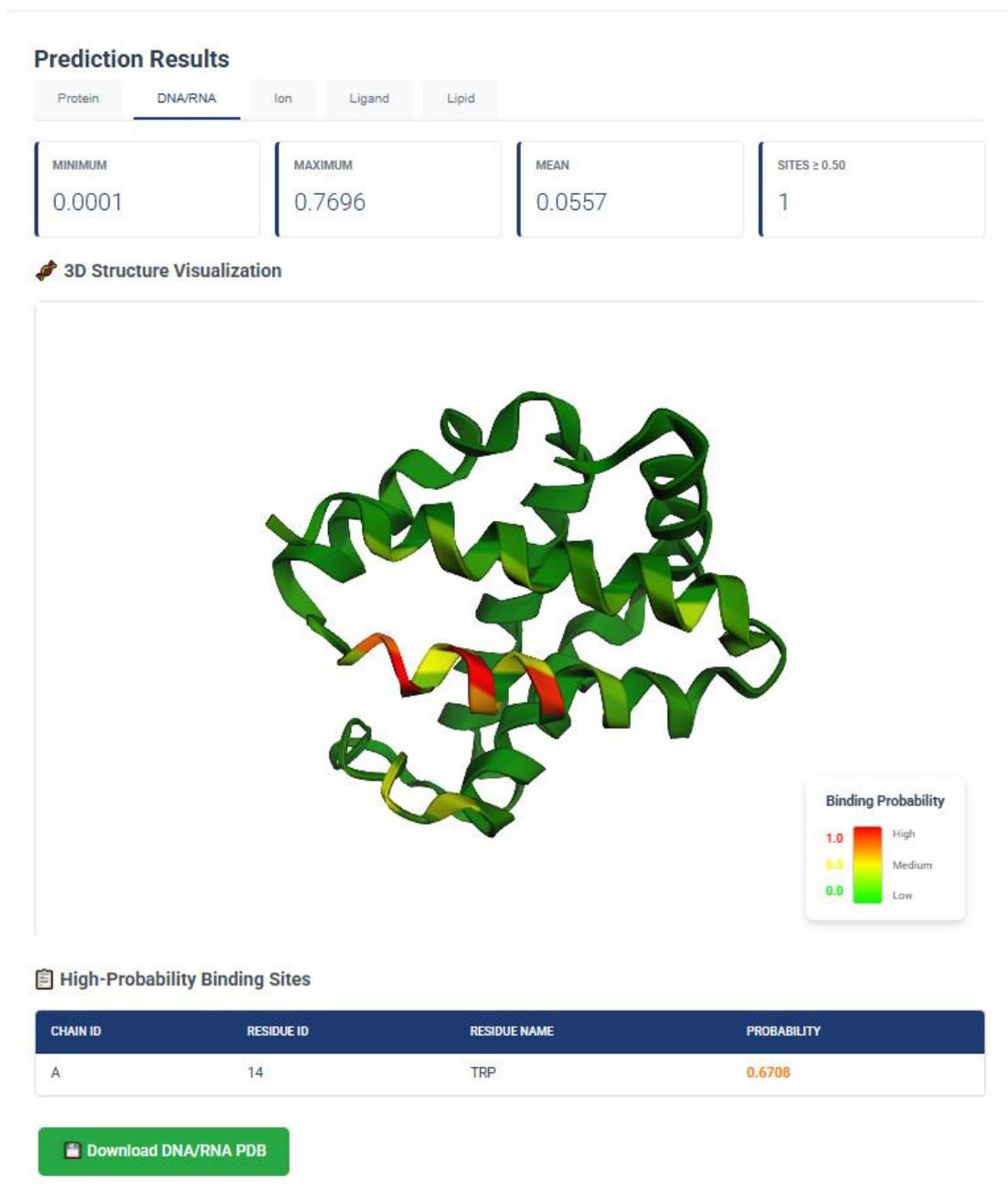

**Figure 7**. Prediction results panel

Users begin by uploading a protein structure file in PDB format via the "Upload PDB File" option. Alternatively, a example sample structure (PDB: 1a01) may be loaded using the "Load Example" button. Following file input, the "Detect Chains" button enables the selection of specific polypeptide chains for analysis. The binding site prediction threshold is then adjustable via a slider labeled "Binding Site Threshold", which spans a continuous range from 0 to 1, with a default preset value of 0.5. Upon configuring these parameters, users initiate the computational prediction by clicking the

"RUN PREDICTION (ALL CLASSES)" button. The system performs simultaneous multi-category binding site prediction within a single inference run. For typical proteins consisting of 200–500 residues, the prediction process is completed within 10–15 seconds, ensuring real-time usability for interactive analysis. The results are subsequently displayed in the "Prediction Results" section of the interface (Figure 7).

The results are organized into several key sections. A summary panel lists positional predictions for interactions with various biomolecular classes, including protein-protein, DNA/RNA, ion, ligan, and lipid binding sites, accompanied by a corresponding legend for interpertation. A dedicated "3D Structure Visualization" panel provides a graphical representation of the protein, employing an intuitive color gradient to represent binding probabilities: green (0.0-0.3), yellow (0.3-0.7), and red (0.7-1.0). Furthermore, a table titled "High-Probability Binding Sites" enumerates specific residues with high confidence scores, detailing the chain identifier, residue number, residue name, and its prediction probability (e.g., Chain A, Residue 14, THR, 0.6788). Finally, to facilitate further analysis, an option to "Download DNA/RNA PDB" is provided. Users can manipulate the 3D structure through standard operations and access detailed residue tables listing high-confidence binding sites, with downloadable PDB files containing probabilities encoded in the B-factor field for compatibility with standard structural biology tools.

## Discussion

This study introduces an edge-aware GAT framework to advance the prediction of protein binding sites across diverse molecular partners. Our approach extends the PeSTo model [22] by integrating multidimensional structural features, incorporating edge-aware graph attention mechanisms, and implementing directinal message passing. These enhancements collectively addres key limitations in capturing fine-grained saptial and anisotropic interactions within protein structures.

We integrated comprehensive structural decriptors—including secondary structure and relative solvent accessibility—with atomic-level geometric features to construct highly informative residue embeddings. This enriched feature representation provides stronger discriminative signals for identifying functional sites compared to using coordinate data alone. Furthermore, we replaced PeSTo's global interaction modeling with sparse, edge-aware graph attention mechanism. By explicityly encoding pairwise Euclidean distances and directional vectors into the attention coefficients (Eq.3), the model prioritizes local spatial dependencies and enforces geometric consistency during message propagation. This allows the network to better capture the complex spatial constraints governing molecular interactions. Finally, the introduction of a learnable query vector for attention-based residue pooling adaptively for attention-based residue pooling adaptively focues aggregation on functionally relevant regions, improving state consolidation for downstream prediction.

As demonstrated in the radar charts (Figure 8a-c), our model achieves a balanced and significant performance gain across all five biomolecular interaction categories compared to the original PeSTo framework. Notably, for protein-protein interactions, accuracy improved from 0.89 to 0.93 and the F1-score increased from 0.69 to 0.77, indicating more precise and balanced predictions. In the challenging lipid-binding category, performance gains were substantial, with the F1-score rising from 0.21 to 0.32 and the Matthews Correlation Coefficient (MCC) improving from 0.20 to 0.46, reflecting a marked reduction in both false positives and false negatives. While a minor fluctuation was observed in the ROC-AUC for ion binding (0.83 vs. PeSTo's 0.86), the overall performance profile remains robust. These results confirm that incorporating edge geometry and directional features, coupled with attentive pooling, yields more informative residue representations and enhances generalization across varied interaction types.

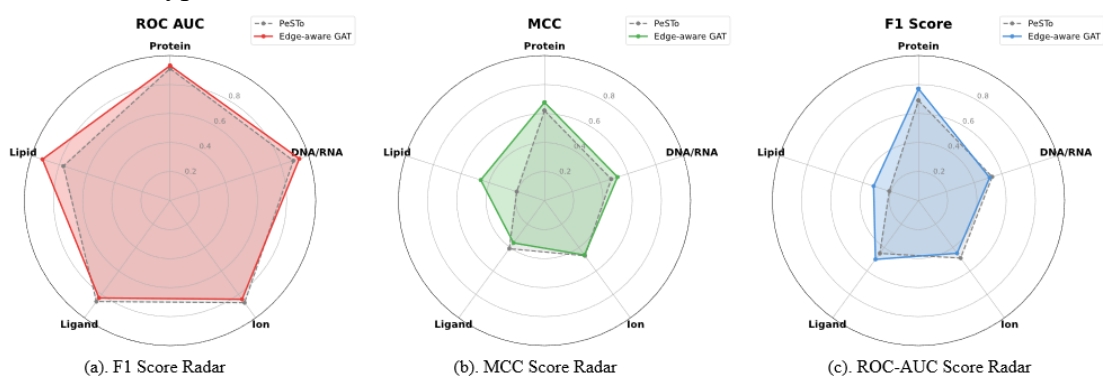

**Figure 8**. Radar charts of F1, MCC, and ROC-AUC score.

The superiority of our framework is further validated through comparison with established baseline methods for protein-protein binding site prediction. As shown in Figure 9, our model achieves a state-of-the-art ROC-AUC of 0.93, outperforming PeSTo (0.91), ScanNet (0.87), MaSIF-site (0.80), Sppider (0.73), and PSIVER (0.64).

This advantage stems from our model's capacity to leverage both atomic-level spatial graphs and geometric edge attributes to precisely model interaction interfaces. In contrast, while PeSTo utilizes atomic geometry, its conservative interaction scoring may limit sensitivity. Methods like ScanNet and MaSIF-site, which rely on specific surface representations or feature schemes, may lose fine-grained atomic details. Sequence-based predictors like Sppider and PSIVER underscore the indispensable value of explicit 3D structural information. Our edge-aware GAT effectively bridges this gap by operating directly on atomic graphs while preserving and propagating critical spatial relationships through an attention mechanism.

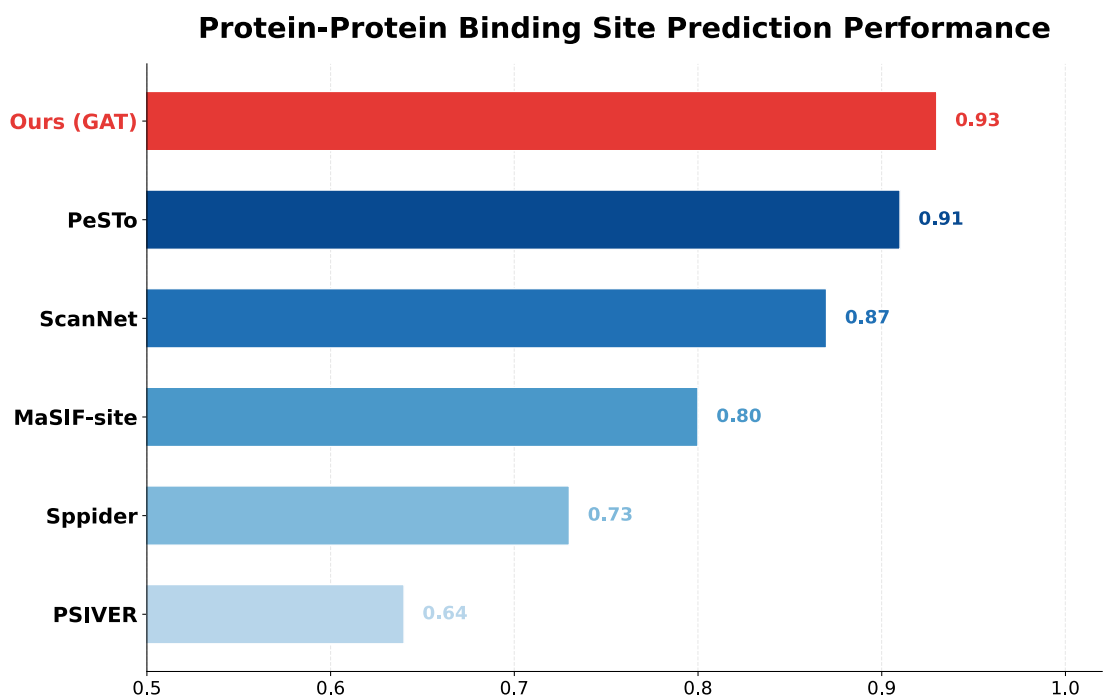

**Figure 9**. ROC-AUC performance comparison of the Edge-aware GAT model against established baseline methods.

Nevertheless, certain limitations persist. Model performance can be influenced by the quality of input protein structures, and reliance on predicted models may introduce uncertainty. Future work will explore the integration of complementary sequence-derived features and evolutionary information to improve robustness against structural variation. Extending the model to predict other functional sites, such as post-translational modification loci, represents a promising direction for broadening its utility. Multi-task learning strategies, as employed in related works, could also be adopted to enhance overall generalization.

In conclusion, by integrating multidimensional structural features, edge-aware graph attention, and directional geometric learning, we have developed a predictive framework that achieves state-of-the-art performance in identifying protein-binding sites. This approach not only improves accuracy but also offers enhanced interpretability into the spatial determinants of molecular recognition, providing a valuable tool for structural bioinformatics and rational drug design.

## Methods

## Data Processing

We employed the same dataset utilized in PesTo, which was constructed from protein structures obtained from the PDB. This ensures direct comparability with the baseline PseTo model and maintains consistency in experimental evaluation.

For each protein subunit, atomic-level coordinates and categorical attributes—including element type, residue type, and atom name—were encoded into feature vectors. Subsequently, spatial topologies were established using a k-nearest neighbor (k-NN) scheme based on Euclidean distances between atoms, with atom pairs within a 5.0 Å cutoff defined as potential contacts.

We maintained the same dataset partitioning as PeSTo, with approximately 70% of the data allocated for training, 15% for validation, and 15% for testing. The test set comprises curated benchmark collections that enable comprehensive evaluation across diverse molecular interaction types, including 53 protein chains from the MaSIF-site benchmark, 230 clustered structures from protein-protein interaction benchmarks, and representative samples from the ScanNet dataset.

Residue-level binding annotations adhered to PeSTo's criteria across five interaction types (proteins/nucleic acids/ligands/ions/lipids), defining binding sites as residues with atoms ≤5 Å from partner-aligning with structural bioinformatics conventions.

While utilizing the same underlying dataset, we extended the feature representation by incorporating additional structural descriptors and implementing our novel edge-aware graph attention framework, as detailed in the following sections.

**Feature Extraction**

**1. Atomic-level Features**

The atomic-level features of the model comprise five components: element type, residue category, atom type, secondary structure and relative solvent accessibility (RSA). The first three features -- element type, residue category and atom type – were formulated as described in [22]. Additionally, we incorporated two features: secondary structure information and RSA. The secondary structure was classified using an 8-category scheme derived from DSSP: α-helix (H), $3_{10}$-helix (G), π-helix (I), extended strand (E), β-bridge (B), turn (T), bend (S), and coil (C) or loop (-), and encoded as 8-dimensional one-hot vector. RSA values were normalized to range [0,1] to represent the degree of residue exposure on the protein surface.

**2. Edge Features**

Edge features between connected atoms includes Eucliean distance (Eq.1) and Unit direction vector (Eq.2). These geometric features enable the model to capture spatial relationships and directional dependencies within the protein structure.

$$d_{ij} = \|x_i - x_j\| \tag{1}$$

$$u_{ij} = \frac{x_i - x_j}{\|x_i - x_j\|} \tag{2}$$

**Model Architecture**

We propose an Edge-aware Graph Attention Network (Edge-aware GAT) that operates on atomic level graphs to predict residue-level binding sites. The architecture consists of three main components (Figure 10)

**1. Feature Embedding Module**

Input features are projected into a 32-dimensional latent space through a stack of three fully connected layers with Exponential Linear Unit (ELU) activation functions. This embedding process generates expressive, compact representations for efficient graph processing.

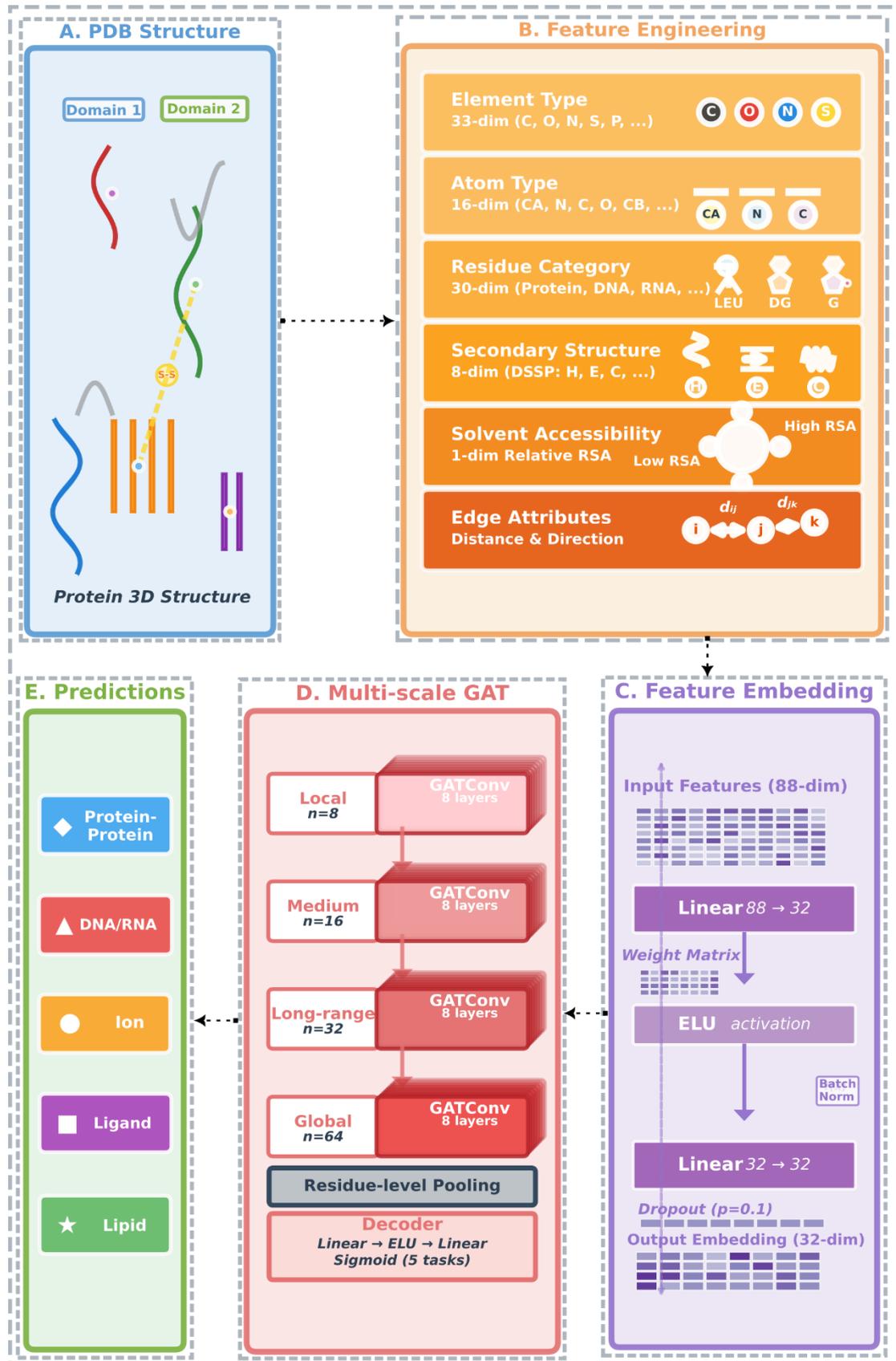

**Figure 10 Overview of the Edge-aware GAT method.** This diagram illustrates the comprehensive architecture of the Edge-aware GAT model for predicting protein binding sites. The workflow proceeds through five sequential stages: (A) **PDB Structure**. Represents the initial protein 3D structure (PDB format), often segmented into domains. (B) **Feature Engineering**. Multidimensional features are extracted. Node (atom-level) features include element type, atom type, residue category, DSSP-derived secondary structure, and relative solvent accessibility (RSA). Edge features comprise pairwise Euclidean distances and unit direction vectors between connected atoms, capturing precise spatial geometry. (C) **Feature Embedding**. The concatenated node features are projected into a compact 32-dimensional latent space via a stack of linear layers with ELU activations and dropout, generating expressive initial embeddings for graph processing. (D) **Multi-scale GAT**. The concatenated node features are projected into a compact 32-dimensional latent space via a stack of linear layers with ELU activations and dropout, generating expressive initial embeddings for graph processing. (E) **Prediction**. Outlines the final output categories for protein binding site predictions, including Protein-Protein, DNA/RNA, Ion, Ligand, and Lipid interactions. The architecture integrates atomic-level spatial graphs with directional edge attributes to model fine-grained, anisotropic molecular interactions for accurate binding sites identification.

## 2. Edge-aware Graph Attention Layers

The core of our model employs 4 graph attention layers that explicitly incorporate edge features into the attention mechanism. For each pair of neighboring nodes i and j, the attention coefficient $a_{ij}$ is computed as:

$$a_{ij} = \frac{\exp\left(LeakyReLU\left(\vec{a}^T[Wh_i \| Wh_i \| e_{ij}]\right)\right)}{\sum_{k \in N(i)} \exp\left(LeakyReLU\left(\vec{a}^T[Wh_i \| Wh_i \| e_{ij}]\right)\right)} \tag{3}$$

where $h_i$, $h_j$ are node features, $W$ is a learnable weight matrix, $e_{ij}$ represents edge features, and ∥ denotes concatenation. The model maintains both scalar state vectors and tensors, which are synchronously updated during message passing:

$$P_i^{(l+1)} = \sum_{j \in N(i)} a_{ij} \cdot d_{ij} \tag{4}$$

This coupled update mechanism ensures geometry-consistent information flow in both scalar and vector spaces.

## 3. Residue-level Pooling and Decoder

An attention-based pooling aggregates atomic features to the residue level. Given a pooling center vector $q_{pool}$, the attention weight $\beta_i$ for residue i is computed as:

$$\beta_i = \frac{exp(q_{pool}^T \cdot q_i)}{\sum_{j=1}^{N} exp(q_{pool}^T \cdot q_j)} \quad (5)$$

The pooled representations are obtained through weighted summation:

$$q_{pooled} = \sum_{i=1}^{N} \beta_i \cdot q_i \quad (6)$$

$$p_{pooled} = \sum_{i=1}^{N} \beta_i \cdot p_i \quad (7)$$

The resulting residue-level features are processed by a multi-layer perceptron decoder that outputs binding probabilities for five molecular classes using multi-label sigmoid activation.

**Training Procedure**

**Loss Function**

To address class imbalance, we employ a weighted multi-label binary cross-entropy loss:

$$\mathcal{L} = -\frac{1}{N}\sum_{i=1}^{N}\sum_{c=1}^{5} \omega_c [y_{ic} \cdot log_\sigma(\hat{y}_{ic}) + (1 - y_{ic}) \cdot log(1 - \sigma(\hat{y}_{ic}))] \quad (8)$$

where σ denotes the sigmoid activation function, and class-specific weights $\omega_c$ are defined as:

$$w_c = \lambda \cdot \frac{1 - r_c}{r_c + \epsilon} \quad (9)$$

Here, $r_c$ denotes the positive sample ratio for class c in the current batch, $\lambda$ is a scaling factor, and $\epsilon$ ensures numerical stability.

**Optimization**

We train the model end-to-end using the Adam optimizer with a fixed learning rate of $1 \times 10^{-5}$ and batch size of 8. Training proceeds for 100 epochs, with model checkpoints saved every 1000 steps. The checkpoint achieving the lowest validation loss is selected for final evaluation.

To enhance training stability, we implement a dummy forward pass for GPU memory preheating and weight initialization, particularly important for handling protein structures of varying sizes and graph topologies.

**Funding**

This work was funded by the National Natural Science Founding of China (62562041, 62162032 and 32260154), and Technology Projects of the Education Department of Jiangxi Province of China (GJJ2201004).